\documentclass[10pt,final,a4paper]{article}
\usepackage[dvips]{graphicx}
\usepackage[utf8]{inputenc}
\usepackage{amssymb,amsmath,array}

\usepackage{algorithm}
\usepackage{algpseudocode}
\usepackage{multirow}

\usepackage{fullpage}

\newcommand{\x}{\mathbf{x}}

\newcommand{\Acal}{\mathcal{A}}
\newcommand{\Bcal}{\mathcal{B}}
\newcommand{\Scal}{\mathcal{S}}
\newcommand{\Fcal}{\mathcal{F}}
\newcommand{\Mcal}{\mathcal{M}}
\newcommand{\argmin}{\operatornamewithlimits{argmin}}

\begin{document}
\title{Very Fast Kernel SVM under Budget Constraints}

\author{David Picard
%
%
\vspace{.3cm}\\
%
ETIS UMR 8051 - ENSEA/Université de Cergy-Pontoise/CNRS \\
6 avenue du Ponceau, F-95000 Cergy, France
}

\maketitle

\begin{abstract}
In this paper we propose a fast online Kernel SVM algorithm under tight budget constraints.
We propose to split the input space using LVQ and train a Kernel SVM in each cluster.
To allow for online training, we propose to limit the size of the support vector set of each cluster using different strategies.
We show in the experiment that our algorithm is able to achieve high accuracy while having a very high number of samples processed per second both in training and in the evaluation.
\end{abstract} 

\section{Introduction and context}

In this paper, we consider Kernel SVM which have been proven to provide very accurate classifiers in wide variety of topics that require binary classification.
In Kernel SVM, a function $k : \Fcal \times \Fcal \mapsto \mathbb{R} $ such that $k(\x_i, \x_j) = \langle \phi(\x_i) , \phi(\x_j) \rangle$ is used to measure the similarity among samples of input space $\Fcal$ instead of using the regular dot product in $\Fcal$.
The classification function $f$ is learned on a training set of labeled samples $\Acal = \{ (\x_i, y_i) \}$:
\begin{align}
	f(\x) = \sum_i \alpha_i k(\x_i, \x)
\end{align}
where $\alpha_i$ are the weights of each training sample.
$f$ is such that its sign should be corresponding to the labels $y_i  \in \{-1, 1\}$ associated with the samples $\x_i$.
As $\phi$ can be non-linear, $f$ is a possibly non linear classification function in $\Fcal$.

To obtain the $\alpha_i$, the following dual problem problem has to be solved:
\begin{align}
	\max_\alpha D(\alpha) &= \sum_i y_i \alpha_i - \frac{1}{2} \sum_{i,j} \alpha_i \alpha_j k(\x_i, \x_j) \\
	\text{s.t. } & \forall i, 0 \leq y_i \alpha_i \leq C
\end{align}

The problem with Kernel SVM is twofold. First, the optimization problem is quadratic with respect to the number of training samples. Second, the decision function computational cost is proportional to the number of non zero $\alpha$ which grows linearly with the size of $\Acal$.
To tackle this problem, most of recent research has been focused on finding explicit approximate mappings $\psi$ such that $\langle \psi(\x_i), \psi(\x_j) \rangle \approx k(\x_i, \x_j)$, that allow to solve the primal problem instead~\cite{le13icml}.
However, such methods are not designed for online scheme and infinite datasets since the kernel approximation often rely on the preprocessing of the whole training set.

In this paper, we focus on Kernel SVM with the ambition to provide an online procedure for infinite datasets.
To achieve this, we propose a budget constraint on the number of support vectors that allows us to design an online learning procedure with a constant cost over time.
To obtain state of the art performances, we propose a hierarchical architecture which splits the input space into subregions, each being assigned a budget constrained Kernel SVM.

The remaining of this paper is as follows: In the next section we present our online update step. In Section~\ref{sec:budget}, we present our strategies for limiting the number of support vectors. In Section~\ref{sec:dc} we present the hierarchical extension. Finally, we present experiments in Section~\ref{sec:exp} before we conclude.

\section{Stochastic Coordinate update}
\label{sec:update}

In the online training scheme, training samples are drawn independently and sequentially from a source.
This means that the size of the learning problem is quadratically increasing with time.
Solving the full quadratic problem anew at each iteration is obviously too costly.
Fortunately, we can build up from the solution of the previous iteration to compute the optimal weights of the current sample.

LaSVM~\cite{bordes05jmlr} is to our knowledge the first of such methods to have been proposed.
At each iteration, the authors compute the weight of the new sample using a quasi-Newton step (which is very efficient since the problem is quadratic). If the weight is non-zero, the samples is added to the \emph{Support Vector set} (SV set), and the SV set is reprocessed (\textit{i.e.}, a quasi-Newton step is applied to all support-vectors, resulting zero-weight samples are removed).
Since the size of the SV set is increasing with time, the reprocess operation tends rapidly to consume all of the computation.

In~\cite{shalev13jmlr}, the authors prove that using the quasi-Newton step alone is sufficient to converge, provided several passes over the entire dataset are made.
In the online scheme, we don't process every sample several time, although some samples might be drawn several times from the source.
Thus, such stochastic update scheme does not converge to the exact solution. However, we found it to give surprisingly good performances in the case of very large datasets.

Base on that, we propose an update procedure in Algorithm~\ref{alg:ocu}.
In this procedure, a new sample $(\x_, y)$ is evaluated against the current SV set.
If it lies inside the margin, then we compute its weight by clamping the result of a quasi-Newton step to the box constraints.
If needed, we then add the sample and the corresponding weight to the SV set.

Remark that most of the computation in this update is performed in the evaluation of $\x$, and is dependent on the size of the current SV set $\Acal$.
As such, we need an efficient way of limiting the size of $\Acal$, as it is explained in the next section.

\begin{algorithm}[H]
\caption{Online Coordinate update}
\label{alg:ocu}
\begin{algorithmic}
\Function{update}{$\Acal = \{ (\x_i, y_i)$ ,$\x$, $y$}
\State $\Scal \leftarrow \Acal$
\State $z \leftarrow \sum_i \alpha_i k(\x_i, \x)$
\If{$y z < 1$}
\State $\Delta\alpha \leftarrow (1 - y_i z)/k(\x_i, \x_i)$
\If{$(\x, \alpha) \in \Scal$}
\State $\alpha \leftarrow y\max(0, \min(C, \alpha+\Delta_\alpha))$
\Else
\State $\Scal \leftarrow \Scal \cup (\x, y\max(0, \min(C, \Delta\alpha))) $
\EndIf
\EndIf
\State \Return $\Scal$
\EndFunction
\end{algorithmic}
\end{algorithm}

\section{Budget constraints}
\label{sec:budget}

The problem of selecting only $n$ support vectors among the whole training set is an assignment problem and as such is very hard to solve.
Most proposed methods to limit the size of the support vector set are heuristic and do not provide strong guaranties neither in term of objective loss nor in terms of test error increase.
However, there are reasonable intuition behind these methods.
In our case of online learning, we want a pruning procedure that remove the least significant support vectors whenever the set becomes too large.
This pruning procedure is presented in Algorithm~\ref{alg:pruning}.

The first method is obviously to remove useless training samples.
Since newer samples can move older ones away from the convex hull of the two classes, it is likely that some optimal weights are to set to 0.
We thus update of weights of the current SV set and keep only samples with non-zero weights (lines 3 to 9).

Then, following the strategy proposed by~\cite{geebelen12tnnls}, we remove samples that are currently misclassified so as to reach the desired size (if possible) based on how far on the wrong side of the margin they are (lines 10 to 15).
The intuition is that such samples are likely to be noise and are deforming the boundary in a very convoluted way.
Not only they require a coefficient for themselves only, but they also provide no generalization which is a hint of overfitting.

Lastly, if the SV set is still over the budget, we remove samples with minimal absolute value weights (lines 16 to 19).
The idea is that these samples are the least contributing to the decision function and thus can be removed as a last resort.
A better choice would be to balance the weight of the samples by the norm of the corresponding columns in the kernel matrix to take into account the influence the sample has on its neighborhood.
However, we didn't found it to be significantly better in practice while being much more costly.

\begin{algorithm}[H]
\caption{SV Set pruning}
\label{alg:pruning}
\begin{algorithmic}[1]
\Function{prune}{$\Acal = \{ (\x_i, \alpha_i) \}$, $n$}
\State $\Scal \leftarrow \emptyset$, $\Bcal \leftarrow \emptyset$
\For{$i=0$ to $\vert \Acal \vert$}
\State \Call{update}{$\Acal$, $\x_i$, $y_i$}
\If{$\alpha_i \neq 0$}
\State $z_i \leftarrow  y_i\sum_n\alpha_nk(\x_i, \x_n)$
\State $\Bcal \leftarrow \Bcal \cup (\x_i, z_i)$
\EndIf
\EndFor
\If{$\vert\Bcal\vert > n$}
\State sort $\Bcal$ by decreasing $z_i$
\State $\theta \leftarrow \min(0, z_n)$
\State $\Bcal \leftarrow \Bcal \setminus \{ (\x_i, z_i) \mid z_i < \theta \} $
\EndIf
\State $\Scal \leftarrow \{(\x_i, \alpha_i) \in \Acal \mid (\x_i, z_i) \in \Bcal \}$
\If{$\vert \Scal  \vert > n$}
\State sort $\Scal$ by decreasing $\vert \alpha_i \vert$
\State $\Scal \leftarrow \Scal \setminus \{ (\x_i, \alpha_i) \mid i > n \} $
\EndIf
\State \Return $\Scal$
\EndFunction
\end{algorithmic}
\end{algorithm}

\section{Divide and Conquer approach}
\label{sec:dc}

Limiting the number of support vectors has the advantage of speeding up the decision function, which has a positive impact on both the training and testing phases, since the online training scheme also uses the evaluation of the decision function to update the weights.
However, it also comes at the price of a simpler boundary which can lead reduced accuracy.

To overcome this problem, we propose to use a divide and conquer approach inspired from ~\cite{hsieh14nips} where the input space is partitioned into $K$ clusters $\mu_k$ and a budget constrained SV set $\Scal_k$ is learned inside each part.
The algorithm has now two parameters, namely the number of clusters $K$ and the budget of each SV set $n$, which make the total storage cost of the model of size $Kn$ support vectors.
However, the evaluation cost is only $\mathcal{O}(K+n)$ since the decision function is only computed in the corresponding cluster.
The full algorithm is detailed in Algorithm~\ref{alg:dcobk}.

To provide a fully online learning scheme, we propose to learn the partition also online using LVQ~\cite{kohonen95lvq}.
While we have not reach this number, we recruit a new cluster centered on the current sample, and associate with it a SV set consisting of the sample and it label as weight (lines 3 to 5).

If $k$ clusters are already present, we assign the current sample to its closest center $k$ (line 7).
We then update the corresponding center using standard LVQ (line 8) and update the SV set $\Scal_k$ using the update online procedure.
We use the pruning procedure to keep the size of $\Scal$ under the constraint if needed.
Although the algorithm is simply to program, our version is publicly available\footnote{download full project at \texttt{https://github.com/davidpicard/dc-bsdca}}.

\begin{algorithm}
\caption{DC-Online Budget Kernel SVM}
\label{alg:dcobk}
\begin{algorithmic}[1]
\Require{Training set $\Acal = \{(\x_i, y_i)\}$}
\Ensure{Local K-SVM set $\Mcal = \{ (\mu_k, \Scal_k)\}$}
\State $\Mcal \leftarrow \emptyset$
\For{$i=0$ to $\vert \Acal \vert$}
\If{$\vert \Mcal \vert < K $}
\State $\mu \leftarrow \x_i,\, \Scal = \{ (\x_i, y_i) \}$
\State $\Mcal \leftarrow \Mcal \cup (\mu, \Scal)$
\Else
\State $k \leftarrow \argmin_m \|\x_i - \mu_m \|$
\State $\mu_k \leftarrow (1-\gamma) \mu_k + \gamma \x_i$
\State \Call{update}{$\Scal_k$, $\x_i$, $y_i$}
\If{$\vert \Scal_k \vert > n$}
\State \Call{prune}{$\Scal_k$, n}
\EndIf
\EndIf
\EndFor
\end{algorithmic}
\end{algorithm}

\section{Experiments}
\label{sec:exp}

We tested Algorithm~\ref{alg:dcobk} on three well known datasets and compared to results reported from~\cite{hsieh14nips}.
Our algorithm was implemented using the JKernelMachines library~\cite{picard13jmlr} and experiments were performed on an Intel CPU running at 2.3GHz.
Hyperparameters (kernel parameters, $C$ and $\gamma$) were chosen by cross-validation on 20\% of the training set.

Results are presented in Table~\ref{tab:res}.
We can see that depending on the choice of the number of clusters $k$ and the budget $n$ for each SV set, we can achieve results close to that of very competitive algorithms.
We also show the throughput of our algorithm in terms of samples processed per second.
As we can see, we almost always can achieve more than 1k samples per second in training and depending on the size of of the model between 1k and 10k samples per seconds in evaluation.

\begin{table}
\centering
\begin{tabular}{|c|c|c|c|c|c|}
\hline
\multirow{2}{*}{Dataset} & \multirow{2}{*}{metric} & \multirow{2}{*}{\cite{hsieh14nips}} & \multicolumn{3}{c|}{DC-Online Budget K-SVM} \\
\cline{4-6}
        &        &           & k=1k,n=1k & k=256,n=64 & k=16,n=64\\
\hline
\hline
CovType & acc & 95.19\% & 94.80\% & 78.95\% & 64.35\% \\
$n_\text{train} = 522k$        & \#/s & -      & 1511 & 1468 &  2821 \\
$n_\text{test} = 58k$        & \#/s & - & 1677 & 7453 & 42.5k \\
\hline
Letter & acc & 95.90\% & 94.82\% & 93.97\% & 83.27\%\\
$n_\text{train} = 12k$        & \#/s & -      & 6066  & 16.5k & 2962 \\
$n_\text{test} = 6k$        & \#/s & -        & 11.1k & 54.0k & 81.1k \\
\hline
USPS & acc & 95.56\%  & 95.71\% & 94.97\% & 85.65\%\\
$n_\text{train} = 7291$        & \#/s & -      & 1101 & 1592 & 988  \\
$n_\text{test} = 2007$        & \#/s & -       & 1948 & 3269 & 26.8k \\
\hline	
\end{tabular}
\caption{Results on 3 different datasets showing the accuracy and the number of samples processed per second for the training step and for the evaluation.}
\label{tab:res}
\end{table}

\section{Conclusion}

In this paper, we presented a Kernel-SVM like algorithm suitable for online learning of very large datasets.
Our algorithm rely on two main ideas, namely a hierarchical approach and a budget constraint.
With the hierarchical approach, we divide the input space using LVQ and train a Kernel SVM for each cluster.
For the budget constraint, we propose strategies to prune the set of support vectors and keep the model complexity reasonable.
We carried out experiments showing our algorithm is capable to achieve comparable results with state of the art classifiers, while having a throughput over 1k samples per second in training and often over 10k samples per second in evaluation.

\bibliographystyle{unsrt}
\bibliography{full}

\begin{thebibliography}{1}

\bibitem{le13icml}
Quoc Le, Tam{\'a}s Sarl{\'o}s, and Alex Smola.
\newblock Fastfood-approximating kernel expansions in loglinear time.
\newblock In {\em Proceedings of the international conference on machine
  learning}, 2013.

\bibitem{bordes05jmlr}
Antoine Bordes, Seyda Ertekin, Jason Weston, and L{\'e}on Bottou.
\newblock Fast kernel classifiers with online and active learning.
\newblock {\em The Journal of Machine Learning Research}, 6:1579--1619, 2005.

\bibitem{shalev13jmlr}
Shai Shalev-Shwartz and Tong Zhang.
\newblock Stochastic dual coordinate ascent methods for regularized loss.
\newblock {\em The Journal of Machine Learning Research}, 14(1):567--599, 2013.

\bibitem{geebelen12tnnls}
Dries Geebelen, Johan~AK Suykens, and Joos Vandewalle.
\newblock Reducing the number of support vectors of svm classifiers using the
  smoothed separable case approximation.
\newblock {\em Neural Networks and Learning Systems, IEEE Transactions on},
  23(4):682--688, 2012.

\bibitem{hsieh14nips}
Cho-Jui Hsieh, Si~Si, and Inderjit~S Dhillon.
\newblock Fast prediction for large-scale kernel machines.
\newblock In {\em Advances in Neural Information Processing Systems}, pages
  3689--3697, 2014.

\bibitem{kohonen95lvq}
Teuvo Kohonen.
\newblock {\em Learning vector quantization}.
\newblock Springer, 1995.

\bibitem{picard13jmlr}
David Picard, Nicolas Thome, and Matthieu Cord.
\newblock Jkernelmachines: a simple framework for kernel machine.
\newblock {\em The Journal of Machine Learning Research}, 14(1):1417--1421,
  2013.

\end{thebibliography}



\end{document}